\documentclass[10pt,twocolumn,letterpaper]{article}

\usepackage[pagenumbers]{cvpr} 

\usepackage[dvipsnames]{xcolor}

\definecolor{cvprblue}{rgb}{0.21,0.49,0.74}
\usepackage[pagebackref,breaklinks,colorlinks,citecolor=cvprblue]{hyperref}
\usepackage{multirow}
\usepackage{cuted}
\usepackage{capt-of}
\usepackage{nth}

\def\paperID{*****} 
\def\confName{CVPR}
\def\confYear{2024}

\title{InstantMesh: Efficient 3D Mesh Generation from a Single Image \\with Sparse-view Large Reconstruction Models}

\author{
Jiale Xu\textsuperscript{\rm 1,2} \quad 
Weihao Cheng\textsuperscript{\rm 1} \quad
Yiming Gao\textsuperscript{\rm 1} \quad
Xintao Wang\textsuperscript{\rm 1}\footnotemark[1] \footnotemark[2] \quad
Shenghua Gao\textsuperscript{\rm 2}\footnotemark[1] \quad
Ying Shan\textsuperscript{\rm 1} \\
\textsuperscript{\rm 1}ARC Lab, Tencent PCG \quad
\textsuperscript{\rm 2}ShanghaiTech University \\
\url{https://github.com/TencentARC/InstantMesh}
}

\begin{document}
\maketitle

\begin{strip}
\vspace{-1.5cm}
\centering
\includegraphics[width=\textwidth]{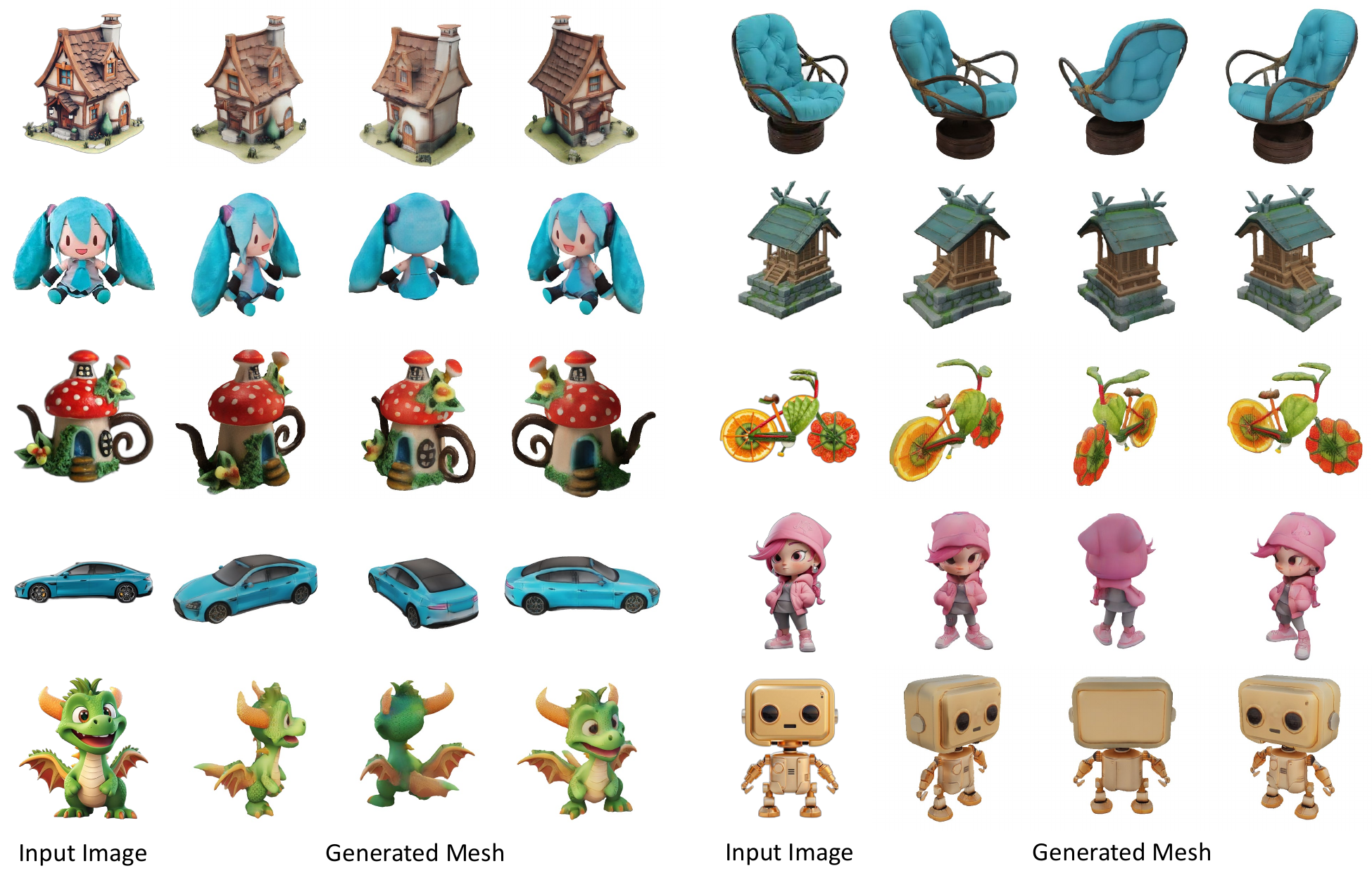}
\captionof{figure}{Given a single image as input, our InstantMesh framework can generate high-quality 3D meshes within 10 seconds.}
\label{fig:teaser}
\end{strip}

\renewcommand{\thefootnote}{\fnsymbol{footnote}}
\footnotetext[1]{Corresponding Authors.}\footnotetext[2]{Project Lead.}

\begin{abstract}

We present InstantMesh, a feed-forward framework for instant 3D mesh generation from a single image, featuring state-of-the-art generation quality and significant training scalability. By synergizing the strengths of an off-the-shelf multiview diffusion model and a sparse-view reconstruction model based on the LRM~\cite{hong2024lrm} architecture, InstantMesh is able to create diverse 3D assets within 10 seconds. To enhance the training efficiency and exploit more geometric supervisions, \eg, depths and normals, we integrate a differentiable iso-surface extraction module into our framework and directly optimize on the mesh representation. Experimental results on public datasets demonstrate that InstantMesh significantly outperforms other latest image-to-3D baselines, both qualitatively and quantitatively. We release all the code, weights, and demo of InstantMesh, with the intention that it can make substantial contributions to the community of 3D generative AI and empower both researchers and content creators.


\end{abstract}    
\section{Introduction}
\label{sec:intro}

Crafting 3D assets from single-view images can facilitate a broad range of applications, 
eg, virtual reality, industrial design, gaming and animation. We have witnessed a revolution on image and video generation with the emergence of large-scale diffusion models~\cite{rombach2022high, saharia2022photorealistic} trained on billion-scale data, which is able to generate vivid and imaginative contents from open-domain prompts. However, duplicating this success on 3D generation presents challenges due to the limited scale and poor annotations of 3D datasets.

To circumvent the problem of lack of 3D data, previous works have explored distilling 2D diffusion priors into 3D representations with a per-scene optimization strategy. 
DreamFusion~\cite{poole2023dreamfusion} proposes score distillation sampling (SDS) which makes a breakthrough in open-world 3D synthesis.
However, SDS with text-to-2D models frequently encounter the multi-face issue, \ie, the ``Janus'' problem. 
To improve 3D consistency, later work~\cite{qian2023magic123} proposes to distill from Zero123~\cite{liu2023zero} which is a novel view generator fine-tuned from Stable Diffusion~\cite{rombach2022high}.
A series of works~\cite{shi2023mvdream, wang2023imagedream, long2023wonder3d, liu2023syncdreamer, voleti2024sv3d} further propose multi-view generation models, thereby the optimization processes can be guided by multiple novel views simultaneously.

2D distillation based methods exhibit strong zero-shot generation capability, but they are time-consuming and not practical for real-world applications. 
With the advent of large-scale open-world 3D datasets~\cite{deitke2023objaverse, deitke2024objaverse}, pioneer works~\cite{hong2024lrm, openlrm, tochilkin2024triposr} demonstrate that 
image tokens can be directly mapped to 3D representations (\eg, triplanes) via a novel large reconstruction model (LRM).
Based on a highly scalable transformer architecture, LRMs point out a promising direction for the fast creation of high-quality 3D assets.
Concurrently, Instant3D~\cite{li2024instantd} proposes a diagram that predicts 3D shapes via an enhanced LRM with multi-view input generated by diffusion models. The method marries LRMs with image generation models, which significantly improves the generalization ability.

LRM-based methods use triplanes as the 3D representation, where novel views are synthesized using an MLP.
Despite the strong geometry and texture representation capability, 
decoding triplanes requires a memory-intensive volume rendering process, which significantly impedes training scales.
Moreover, the expensive computational overhead makes it challenging to utilize high-resolution RGB and geometric information (\eg, depths and normals) for supervision. To boost the training efficiency, recent works seek to utilize Gaussians~\cite{kerbl20233d} as the 3D representation, which is effective for rendering but not suitable for geometric modeling. Several concurrent works~\cite{zheng2024mvd, wang2024crm} opt to apply supervisions on the mesh representation directly using differentiable surface optimization techniques~\cite{shen2021deep, shen2023flexible}. However, they adopt CNN-based architectures, which limit their flexibility to deal with varying input viewpoints and training scalability on larger datasets that may be available in the future.

In this work, we present InstantMesh, a feed-forward framework for high-quality 3D mesh generation from a single image. Given an input image, InstantMesh first generates 3D consistent multi-view images with a multi-view diffusion model, and then utilizes a sparse-view large reconstruction model to predict a 3D mesh directly, where the whole process can be accomplished in seconds. By integrating a differentiable iso-surface extraction module, our reconstruction model applies geometric supervisions on the mesh surface directly, enabling satisfying training efficiency and mesh generation quality. Building upon an LRM-based architecture, our model offers superior training scalability to large-scale datasets. Experimental results demonstrate that InstantMesh outperforms other latest image-to-3D approaches significantly. We hope that InstantMesh can serve as a powerful image-to-3D foundation model and make substantial contributions to the field of 3D generative AI.

\section{Related Work}
\label{sec:related}

\noindent\textbf{Image-to-3D.}
Early attempts on image-to-3D mainly focus on the single-view reconstruction task~\cite{zhou20213d, wang2018pixel2mesh, pan2019deep, chen2020bsp, mescheder2019occupancy, niemeyer2020differentiable}. With the rise of diffusion models, pioneer works have investigated image-conditioned 3D generative modeling on various representations, \eg, point clouds~\cite{nichol2022point, melas2023pc2, wu2023sketch, zhou20213d, tyszkiewicz2023gecco}, meshes~\cite{liu2023meshdiffusion, alliegro2023polydiff}, SDF grids~\cite{zheng2023locally, cheng2023sdfusion, chou2023diffusion, shim2023diffusion} and neural fields~\cite{gupta20233dgen, muller2023diffrf, zhang20233dshape2vecset, jun2023shap, wang2023rodin}. Despite the promising progress these methods have made, they are hard to generalize to open-world objects due to the limited scale of training data.

\begin{figure*}[t]
\centering
\includegraphics[width=\textwidth]{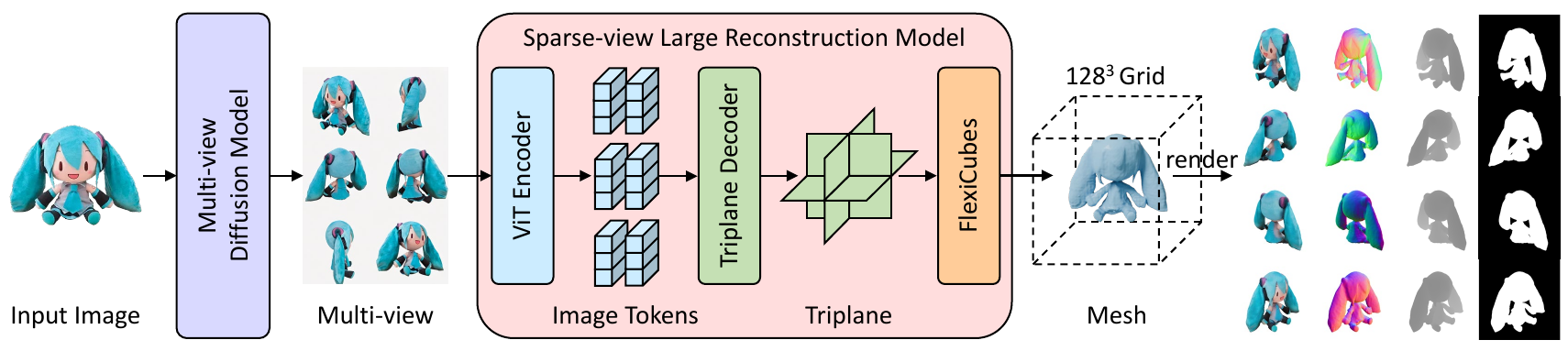}
\caption{The overview of our InstantMesh framework. Given an input image, we first utilize a multi-view diffusion model to synthesize 6 novel views at fixed camera poses. Then we feed the generated multi-view images into a transformer-based sparse-view large reconstruction model to reconstruct a high-quality 3D mesh. The whole image-to-3D generation process takes only around 10 seconds. By integrating an iso-surface extraction module, \ie, FlexiCubes, we can render the 3D geometry efficiently and apply geometric supervisions like depths and normals directly on the mesh representation to enhance the results.}
\label{fig:pipeline}
\end{figure*}

The advent of powerful text-to-image diffusion models~\cite{saharia2022photorealistic, rombach2022high} inspires the idea of distilling 2D diffusion priors into 3D neural radiance fields with a per-scene optimization strategy. The score distillation sampling (SDS) proposed by DreamFusion~\cite{poole2023dreamfusion} exhibits superior performance on zero-shot text-to-3D synthesis and outperforms CLIP-guided alternatives~\cite{radford2021learning, jain2022zero, xu2023dream3d} significantly. However, SDS-based methods~\cite{wang2023score, lin2023magic3d, chen2023fantasia3d, wang2024prolificdreamer} frequently encounter the multi-face issue, also known as the ``Janus'' problem. Zero123~\cite{liu2023zero} demonstrates that Stable Diffusion can be fine-tuned to synthesize novel views by conditioning on relative camera poses. Leveraging the novel view guidance provided by Zero123, recent image-to-3D methods~\cite{liu2024one, qian2023magic123, xu2023neurallift} show improved 3D consistency and can generate plausible shapes from open-domain images.

\noindent\textbf{Multi-view Diffusion Models.}
To address the inconsistency among multiple generated views of Zero123, some works~\cite{liu2023syncdreamer, long2023wonder3d, shi2023zero123++, wang2023imagedream} try to fine-tune 2D diffusion models to synthesize multiple views for the same object simultaneously. With 3D consistent multi-view images, various techniques can be applied to obtain the 3D object, \eg, SDS optimization~\cite{wang2023imagedream}, neural surface reconstruction methods~\cite{liu2023syncdreamer, long2023wonder3d}, multi-view-conditioned 3D diffusion models~\cite{liu2023one}. To further enhance the generalization capability and multi-view consistency, some recent works\cite{voleti2024sv3d, chen2024v3d, han2024vfusion3d, zuo2024videomv} exploit the temporal priors in video diffusion models for multi-view generation.



\noindent\textbf{Large Reconstruction Models.}
The availability of large-scale 3D datasets~\cite{deitke2023objaverse, deitke2024objaverse} enables training highly generalizable reconstruction models for feed-forward image-to-3D creation. Large Reconstruction Model~\cite{hong2024lrm, wang2024pflrm, li2024instantd, xu2024dmvd} (LRM) demonstrates that the transformer backbone can effectively map image tokens to implicit 3D triplanes with multi-view supervision. Instant3D~\cite{li2024instantd} further extends LRM to sparse-view input, significantly boosting the reconstruction quality. By combining with multi-view diffusion models, Instant3D can achieve highly generalizable and high-quality single-image to 3D generation. Inspired by Instant3D, LGM~\cite{tang2024lgm} and GRM~\cite{xu2024grm} replace the triplane NeRF~\cite{mildenhall2021nerf} representation with 3D Gaussians~\cite{kerbl20233d} to enjoy its superior rendering efficiency and circumvent the need for memory-intensive volume rendering process. However, Gaussians fall short on explicit geometry modeling and high-quality surface extraction. Given the success of neural mesh optimization methods~\cite{shen2021deep, shen2023flexible}, concurrent works MVD$^2$~\cite{zheng2024mvd} and CRM~\cite{wang2024crm} opt to optimize on the mesh representation directly for efficient training and high-quality geometry and texture modeling. Different from their convolutional network architecture, our model is built upon LRM and opts for a purely transformer-based architecture, offering superior flexibility and training scalability.


\section{InstantMesh}

The architecture of InstantMesh is similar to Instant3D~\cite{li2024instantd}, consisting of a multi-view diffusion model $G_M$ and a sparse-view large reconstruction model $G_R$. Given an input image $I$, $G_M$ generates 3D consistent multi-view images from $I$, which are fed into $G_R$ to reconstruct a high-quality 3D mesh. We now introduce our technical improvements on data preparation, model architecture and training strategies.

\subsection{Multi-view Diffusion Model}
Technically, our sparse-view reconstruction model accepts free-viewpoint images as input, so we can integrate arbitrary multi-view generation model into our framework,
\eg, MVDream~\cite{shi2023mvdream}, ImageDream~\cite{wang2023imagedream}, SyncDreamer~\cite{liu2023syncdreamer}, SPAD~\cite{kant2024spad} and SV3D~\cite{voleti2024sv3d}, to achieve both text-to-3D and image-to-3D assets creation. We opt for Zero123++~\cite{shi2023zero123++} due to its reliable multi-view consistency and tailored viewpoint distribution that covers both the upper and lower parts of a 3D object.

\noindent\textbf{White-background Fine-tuning.}
Given an input image, Zero123++ generates a $960\times 640$ gray-background image presenting 6 multi-view images in a $3\times 2$ grid. In practice, we notice that the generated background is not consistent across different image areas and varies in RGB values, leading to floaters and cloud-like artifacts in the reconstruction results. And LRMs are often trained on white-background images too. To remove the gray background, we need to utilize third-party libraries or models that cannot guarantee the segmentation consistency among multiple views. Therefore, we opt to fine-tune Zero123++ to synthesize consistent white-background images, ensuring the stability of the latter sparse-view reconstruction procedure.

\noindent\textbf{Data Preparation and Fine-tuning Details.}
We prepare the fine-tuning data following the camera distribution of Zero123++. Specifically, for each 3D model in the LVIS subset of Objaverse~\cite{deitke2023objaverse}, we render a query image and 6 target images, all in white backgrounds. The azimuth, elevation and camera distance of the query image is randomly sampled from a pre-defined range. The poses of the 6 target images consist of interleaving absolute elevations of $20^{\circ}$ and $-10^{\circ}$, combined with azimuths relative to the query image that start at $30^{\circ}$ and increase by $60^{\circ}$ for each pose.

During fine-tuning, we use the query image as the condition and stitch the 6 target images into a $3\times 2$ grid for denoising. Following Zero123++, we adopt the linear noise schedule and $v$-prediction loss. We also randomly resize the conditional image to make the model adapt to various input resolutions and generate clear images. Since the goal of fine-tuning is a simple replacement of background color, it converges extremely fast. Specifically, we fine-tune the UNet for 1000 steps with a learning rate of $1.0\times 10^{-5}$ and a batch size of 48. The fine-tuned model can fully preserve the generation capability of Zero123++ and produce white-background images consistently.

\subsection{Sparse-view Large Reconstruction Model}
We present the details of the sparse-view reconstruction model $G_R$ that predicts meshes given generated multi-view images. The architecture of $G_R$ is modified and enhanced from Instant3D~\cite{li2024instantd}.



\noindent\textbf{Data Preparation.}
Our training dataset is composed of multi-view images rendered from the Objaverse~\cite{deitke2023objaverse} dataset. Specifically, we render $512\times 512$ images, depths and normals from 32 random viewpoints for each object in the dataset. Besides, we use a filtered high-quality subset to train our model. The filtering goal is to remove objects that satisfy any of the following criteria: (i) objects without texture maps, (ii) objects with rendered images occupying less than $10\%$ of the view from any angle, (iii) including multiple separate objects, (iv) objects with no caption information provided by the Cap3D dataset, and (v) low-quality objects. The classification of ``low-quality" objects is determined based on the presence of tags such as ``lowpoly" and its variants (e.g., 
``low\_poly") in the metadata. Specifically, by applying our filtering criteria, we curated approximately 270k high-quality instances from the initial pool of 800k objects in the Objaverse dataset.

\noindent\textbf{Input Views and Resolution.}
During training, we randomly select a subset of 6 images as input and another 4 images as supervision for each object. To be consistent with the output resolution of Zero123++, all the input images are resized to $320\times 320$. During inference, we feed the 6 images generated by Zero123++ as the input of the reconstruction model, whose camera poses are fixed. To be noted, our transformer-based architecture makes it natural to utilize varying number of input views, thus it is practical to use less input views for reconstruction, which can alleviate the multi-view inconsistency issue in some cases.

\noindent\textbf{Mesh as 3D Representation.}
Previous LRM-based methods output triplanes that require volume rendering to synthesize images. During training, volume rendering is memory expensive that hinders the use of high-resolution images and normals for supervision.
To enhance the training efficiency and reconstruction quality, we integrate a differentiable iso-surface extraction module, \ie, FlexiCubes~\cite{shen2023flexible}, into our reconstruction model. Thanks to the efficient mesh rasterization, we can use full-resolution images and additional geometric information for supervision, \eg, depths and normals, without cropping them into patches. Applying these geometric supervisions leads to smoother mesh outputs compared to the meshes extracted from the triplane NeRF. Besides, using mesh representation can also bring convenience to applying additional post-processing steps to enhance the results, such as SDS optimization~\cite{lin2023magic3d, chen2023fantasia3d} or texture baking~\cite{liu2024one}. We leave it as a future work.

Different from the single-view LRM, our reconstruction model takes 6 views as input, requiring more memory for the cross-attention between the triplane tokens and image tokens. We notice that training such a large-scale transformer from scratch requires a significant period of time. For faster convergence, we initialize our model using the pre-trained weights of OpenLRM~\cite{openlrm}, an open-source implementation of LRM. We adopt a two-stage training strategy as described below.

\noindent\textbf{Stage 1: Training on NeRF.}
In the first stage, we train on the triplane NeRF representation and reuse the prior knowledge of the pre-trained OpenLRM. To enable multi-view input, we add AdaLN camera pose modulation layers in the ViT image encoder to make the output image tokens pose-aware following Instant3D, and remove the source camera modulation layers in the triplane decoder of LRM. We adopt both image loss and mask loss in this training stage:
\begin{equation}
\begin{aligned}
\mathcal{L}_1 &= \sum_i \left\|\hat{I}_i-I_i^{gt}\right\|_2^2 \\
&+ \lambda_{\text{lpips}} \sum_i \mathcal{L}_{\text{lpips}}\left(\hat{I}_i, I_i^{gt}\right) \\
&+ \lambda_{\text{mask}} \sum_i \left\|\hat{M}_i-M_i^{gt}\right\|_2^2,
\end{aligned}
\end{equation}
where $\hat{I}_i$, $I_i^{gt}$, $\hat{M}_i$ and $M_i^{gt}$ denote the rendered images, ground truth images, rendered mask, and ground truth masks of the $i$-th view, respectively. During training, we set $\lambda_{\text{lpips}}=2.0, \lambda_{\text{mask}} = 1.0$, and use a learning rate of $4.0\times 10^{-4}$ cosine-annealed to $4.0\times 10^{-5}$. To enable high-resolution training, our model renders $192\times 192$ patches which are supervised by cropped ground truth patches ranging from $192\times 192$ to $512\times 512$.

\noindent\textbf{Stage 2: Training on Mesh.}
In the second stage, we switch to the mesh representation for efficient training and applying additional geometric supervisions. We integrate FlexiCubes~\cite{shen2023flexible} into our reconstruction model to extract mesh surface from the triplane implicit fields. The original triplane NeRF renderer consists of a density MLP and a color MLP, we reuse the density MLP to predict SDF instead, and add two additional MLPs to predict the deformation and weights required by FlexiCubes. 

For a density field $f(\mathbf{x})=d, \mathbf{x}\in\mathbb{R}^3$, points inside the object have larger values and points outside the object have smaller values, while an SDF field $g(\mathbf{x})=s$ is just the opposite. Therefore, we initialize the weight $\mathbf{w}\in\mathbb{R}^{C}$ and bias $b\in\mathbb{R}$ of the last SDF MLP layer as follows:
\begin{equation}
\begin{aligned}
    \mathbf{w} &= -\mathbf{w}_d, \\
    b &= \tau - b_d,
\end{aligned}
\end{equation}
where $\mathbf{w}_d\in\mathbb{R}^{C}$ and $b_d\in\mathbb{R}$ are the weight and bias of the original density MLP's last layer, and $\tau$ denotes the iso-surface threshold used for density fields. Denoting the input feature of the last MLP layer as $\mathbf{f}\in\mathbb{R}^{C}$, we have 
\begin{equation}
\begin{aligned}
    s &= \mathbf{w}\cdot\mathbf{f}+b \\
    &= (-\mathbf{w}_d)\cdot\mathbf{f} + (\tau-b_d) \\
    &= -(\mathbf{w}_d\cdot\mathbf{f} + b_d - \tau) \\
    &= -(d-\tau),
\end{aligned}
\end{equation}

With such an initialization, we reverse the ``direction" of density field to match the SDF direction and ensure that the iso-surface boundary lies at the 0 level-set of the SDF field at the beginning. We empirically find that this initialization benefits the training stability and convergence speed of FlexiCubes. The loss function of the second stage is:
\begin{equation}
\begin{aligned}
\mathcal{L}_2 = \mathcal{L}_1 &+ \lambda_{\text{depth}} \sum_i M^{gt}\otimes\left\|\hat{D}_i-D_i^{gt}\right\|_1 \\
&+ \lambda_{\text{normal}} \sum_i M^{gt}\otimes\left(1 - \hat{N}_i\cdot N_i^{gt}\right) \\
&+ \lambda_{\text{reg}}\mathcal{L}_{\text{reg}},
\end{aligned}
\end{equation}
where $\hat{D}_i$, $D_i^{gt}$, $\hat{N}_i$ and $ N_i^{gt}$ denote the rendered depth, ground truth depth, rendered normal and ground truth normal of the $i$-th view, respectively. $\otimes$ denotes element-wise production, and $\mathcal{L}_{\text{reg}}$ denotes the regularization terms of FlexiCubes. During training, we set $\lambda_{\text{depth}}=0.5, \lambda_{\text{normal}}=0.2, \lambda_{\text{reg}}=0.01$, and use a learning rate of $4.0\times10^{-5}$ cosine-annealed to 0. We train our model on 8 NVIDIA H800 GPUs in both stages.

\noindent\textbf{Camera Augmentation and Perturbation.}
Different from view-space reconstruction models~\cite{openlrm, tochilkin2024triposr, tang2024lgm, zou2023triplane}, our model reconstruct 3D objects in a canonical world space where the $z$-axis aligns with the anti-gravity direction. To further improve the robustness on the scale and orientation of 3D objects, we perform random rotation and scaling on the input multi-view camera poses. Considering that the multi-view images generated by Zero123++ may be inconsistent with their pre-defined camera poses, we also add random noise to the camera parameters before feeding them into the ViT image encoder.

\begin{table}[t]
\small
\renewcommand{\tabcolsep}{3.3mm}
\caption{Details of sparse-view reconstruction model variants.}
\label{tab:variants}
\centering
\begin{tabular}{@{}l|c|c|c|c}
\toprule
\multirow{2}{*}{Parameter} & \multicolumn{2}{c}{InstantNeRF} & \multicolumn{2}{c}{InstantMesh} \\
\cmidrule(r){2-3} \cmidrule(r){4-5}
& base & large & base & large \\
\midrule
Representation & NeRF & NeRF & Mesh & Mesh \\
Input views & 6 & 6 & 6 & 6 \\
Transformer dim & 1024 & 1024 & 1024 & 1024 \\
Transformer layers & 12 & 16 & 12 & 16 \\
Triplane size & 64 & 64 & 64 & 64 \\
Triplane dim & 40 & 80 & 40 & 80 \\
Samples per ray & 96 & 128 & - & - \\
Grid size & - & - & 128 & 128 \\
Input size & 320 & 320 & 320 & 320 \\
Render size & 192 & 192 & 512 & 512 \\
\bottomrule
\end{tabular}
\end{table}

\noindent\textbf{Model Variants.}
In this work, we provide 4 variants of the sparse-view reconstruction model, two from Stage 1 and two from Stage 2. We name each model according to its 3D representation (``NeRF" or ``Mesh") and the scale of parameters (``base" or ``large"). The details of each model are shown in Table~\ref{tab:variants}. Considering that different 3D presentations and model scales can bring convenience to different application scenarios, we release the weights of all the 4 models. We believe our work can serve as a powerful image-to-3D foundation model and facilitate future research on 3D generative AI.

\section{Experiments}

In this section, we conduct experiments to compare our InstantMesh with existing state-of-the-art image-to-3D baseline methods quantitatively and qualitatively. 

\subsection{Experimental Settings}

\noindent\textbf{Datasets.} 
We evaluate the quantitative performance using two public datasets, \ie, Google Scanned Objects (GSO)~\cite{downs2022google} and OmniObject3D (Omni3D)~\cite{wu2023omniobject3d}. GSO contains around 1K objects, from which we randomly pick out 300 objects as the evaluation set. For Omni3D, we select 28 common categories and then pick out the first 5 objects from each category for a total of 130 objects (some categories have less than 5 objects) as the evaluation set.

To evaluate the 2D visual quality of the generated 3D meshes, we create two image evaluation sets for both GSO and Omni3D. Specifically, we render 21 images of each object in an orbiting trajectory with uniform azimuths and varying elevations in $\{30^{\circ}, 0^{\circ}, -30^{\circ}\}$. As Omni3D also includes benchmark views randomly sampled on the top semi-sphere of an object, we pick 16 views randomly and create an additional image evaluation set for Omni3D.

\noindent\textbf{Baselines.} 
We compare the proposed InstantMesh with 4 baselines: (i) TripoSR \cite{tochilkin2024triposr}: an open-source LRM implementation showing the best single-view reconstruction performance so far; (ii) LGM \cite{tang2024lgm}: a unet-based Large Gaussian Model that reconstructs Gaussians from generated multi-view images; (iii) CRM \cite{wang2024crm}: a unet-based Convolutional Reconstruction Model that reconstructs 3D meshes from generated multi-view images and canonical coordinate maps (CCMs). (iv) SV3D \cite{voleti2024sv3d}: an image-conditioned diffusion model based on Stable Video Diffusion~\cite{blattmann2023stable} that generates an orbital video of an object, we only evaluate it on the novel view synthesis task since generating 3D meshes from its output is not straight-forward.

\noindent\textbf{Metrics.} 
We evaluate both the 2D visual quality and 3D geometric quality of the generated assets. For 2D visual evaluation, we render novel views from the generated 3D mesh and compare them with the ground truth views, and adopt PSNR, SSIM, and LPIPS as the metrics. For 3D geometric evaluation, we first align the coordinate system of the generated meshes with the ground truth meshes, and then reposition and re-scale all meshes into a cube of size $[-1, 1]^3$. We report Chamfer Distance (CD) and F-Score (FS) with a threshold of $0.2$, which are computed by sampling 16K points from the surface uniformly.

\subsection{Main Results}

\noindent\textbf{Quantitative Results.}
We report the quantitative results on different evaluation sets in Table \ref{tab:gso}, \ref{tab:omni3d}, and \ref{tab:omni3dbm}, respectively. For each metric, we highlight the top three results among all methods, and a deeper color indicates a better result. For our method, we report the results of using different sparse-view reconstruction model variants (\ie, ``NeRF" and ``Mesh").

\begin{figure*}[t]
\centering
\includegraphics[width=\textwidth]{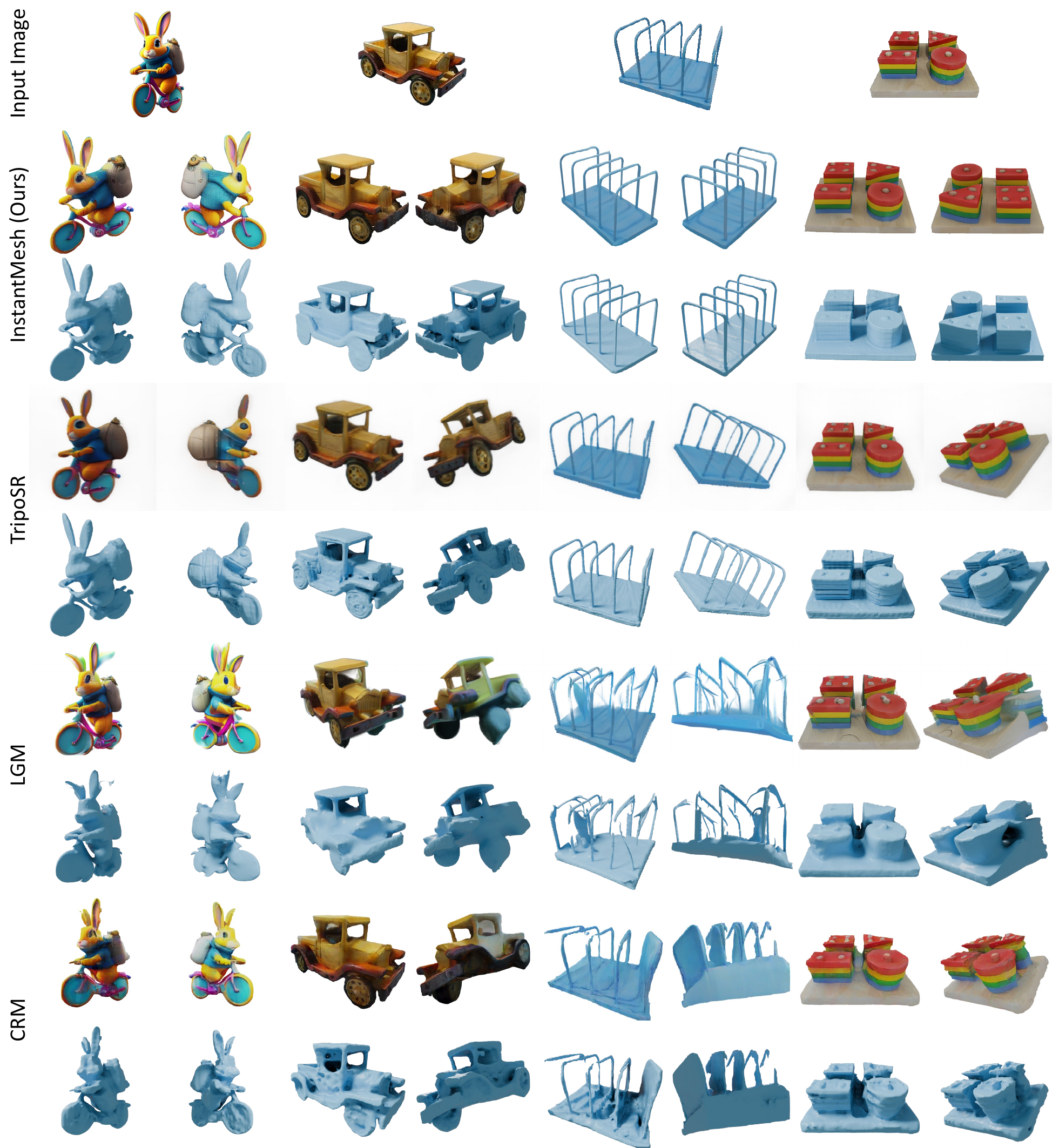}
\caption{
The 3D meshes generated by InstantMesh demonstrate significantly better geometry and texture compared to the other baselines. The results of InstantMesh are rendered at a fixed elevation of $20^{\circ}$, while the results of other methods are rendered at a fixed elevation of $0^{\circ}$ since they reconstruct objects in the view space.}
\label{fig:comparison}
\end{figure*}

From the 2D novel view synthesis metrics, we can observe that InstantMesh outperforms the baselines on SSIM and LPIPS significantly, indicating that its generation results have the best perceptually viewing quality. As Figure \ref{fig:comparison} shows, InstantMesh demonstrates plausible appearances, whereas the baselines frequently exhibit distortions in novel views. We can also observe that the PSNR of InstantMesh is slightly lower than the best baseline, suggesting that the novel views are less faithful to the ground truth at pixel level since they are ``dreamed" by the multi-view diffusion model. However, we argue that the perceptual quality is more important than faithfulness, as the ``true novel views'' should be unknown and have multiple possibilities given a single image as reference.

As for the 3D geometric metrics, InstantMesh outperforms the baselines on both CD and FS significantly, which indicates a higher fidelity of the generated shapes. From Figure \ref{fig:comparison}, we can observe that InstantMesh presents the most reliable geometries among all methods. Benefiting from the scalable architecture and tailored training strategies, InstantMesh achieves the state-of-the-art image-to-3D performance.

\newcommand{\mrka}[1]{{\colorbox{red!30}{#1}}}
\newcommand{\mrkb}[1]{{\colorbox{red!20}{#1}}}
\newcommand{\mrkc}[1]{{\colorbox{red!10}{#1}}}

\begin{table}[t]
  \small
  \renewcommand{\tabcolsep}{1.5mm}
  \caption{Quantitative results on Google Scanned Objects (GSO) orbiting views.}
  \centering
  \begin{tabular}{@{}l|c|c|c|c|c}
    \toprule
     Method & PSNR $\uparrow$ & SSIM $\uparrow$ & LPIPS $\downarrow$ & CD $\downarrow$ & FS $\uparrow$ \\
    \midrule
    TripoSR & \mrka{23.373} & 0.868 & 0.213 & \mrkc{0.217} & \mrkc{0.843} \\
    LGM & 21.538 & 0.871 & 0.216 & 0.345 & 0.671 \\
    CRM & 22.195 & \mrkc{0.891} & \mrkc{0.150} & 0.252 & 0.787 \\
    SV3D & 22.098 & 0.861 & 0.201 & - & - \\
    Ours (NeRF) & \mrkb{23.141} & \mrka{0.898} & \mrka{0.119} & \mrka{0.177} & \mrka{0.882}  \\
    Ours (Mesh) & \mrkc{22.794} & \mrkb{0.897} & \mrkb{0.120} & \mrkb{0.180} & \mrkb{0.880} \\
    \bottomrule
  \end{tabular}
  \label{tab:gso}
\end{table}

\begin{table}[t]
  \small
  \renewcommand{\tabcolsep}{1.5mm}
  \caption{Quantitative results on OmniObject3D (Omni3D) orbiting views.}
  \centering
  \begin{tabular}{@{}l|c|c|c|c|c}
    \toprule
     Method & PSNR $\uparrow$ & SSIM $\uparrow$ & LPIPS $\downarrow$ & CD $\downarrow$ & FS $\uparrow$ \\
    \midrule
    TripoSR & \mrkb{21.996} & 0.877 & 0.198 & \mrkc{0.245} & \mrkc{0.811} \\
    LGM & 20.434 & 0.864 & 0.226 & 0.382 & 0.635 \\
    CRM & 21.630 & \mrkc{0.892} & \mrkc{0.147} & 0.246 & 0.802 \\
    SV3D & 21.510 & 0.866 & 0.186 & - & - \\
    Ours (NeRF) & \mrka{22.635} & \mrka{0.903} & \mrka{0.110} & \mrka{0.199} & \mrka{0.869} \\
    Ours (Mesh) & \mrkc{21.954} & \mrkb{0.901} & \mrkb{0.112} & \mrkb{0.203} & \mrkb{0.864} \\
    \bottomrule
  \end{tabular}
  \label{tab:omni3d}
\end{table}

\begin{table}[t]
  \small
  \renewcommand{\tabcolsep}{1.5mm}
  \caption{Quantitative results on OmniObject3D (Omni3D) benchmark views.}
  \centering
  \begin{tabular}{@{}l|c|c|c|c|c}
    \toprule
     Method & PSNR $\uparrow$ & SSIM $\uparrow$ & LPIPS $\downarrow$ & CD $\downarrow$ & FS $\uparrow$ \\
    \midrule
    TripoSR & \mrkb{19.977} & 0.859 & 0.206 & \mrkc{0.221} & \mrkc{0.847} \\
    LGM & 18.665 & 0.832 & 0.250 & 0.356 & 0.653 \\
    CRM & 19.422 & \mrkc{0.865} & \mrkc{0.172} & 0.274 & 0.778 \\
    SV3D & \mrka{20.294} & 0.853 & 0.176 & - & - \\
    Ours (NeRF) & \mrkc{19.752} & \mrka{0.869} & \mrkb{0.150} & \mrkb{0.206} & \mrkb{0.863} \\
    Ours (Mesh) & 19.552 & \mrkb{0.868} & \mrka{0.150} & \mrka{0.204} & \mrka{0.866} \\
    \bottomrule
  \end{tabular}
  \label{tab:omni3dbm}
\end{table}

\noindent\textbf{Qualitative Results.}
To compare our InstantMesh with other baselines qualitatively, we select two images from the GSO evaluation set and two images from Internet, and obtain the image-to-3D generation results. For each generated mesh, we visualize both the textured renderings (upper) and pure geometry (lower) from two different viewpoints. We use the ``Mesh" variant of sparse-view reconstruction model to generate our results. 

As depicted in Figure \ref{fig:comparison}, the generated 3D meshes of InstantMesh present significantly more plausible geometry and appearance. TripoSR can generate satisfactory results from images that have a similar style to the Objaverse dataset, but it lacks the imagination ability and tends to generate degraded geometry and textures on the back when the input image is more free-style (Figure~\ref{fig:comparison}, \nth{3} row, \nth{1} column). Thanks to the high-resolution supervision, InstantMesh can also generate sharper textures compared to TripoSR. LGM and CRM share a similar framework to ours by combining a multi-view diffusion model with a sparse-view reconstruction model, thus they also enjoy the imagination ability. However, LGM exhibits distortions and obvious multi-view inconsistency, while CRM has difficulty in generating smooth surfaces.

\begin{figure*}[t]
\centering
\includegraphics[width=\textwidth]{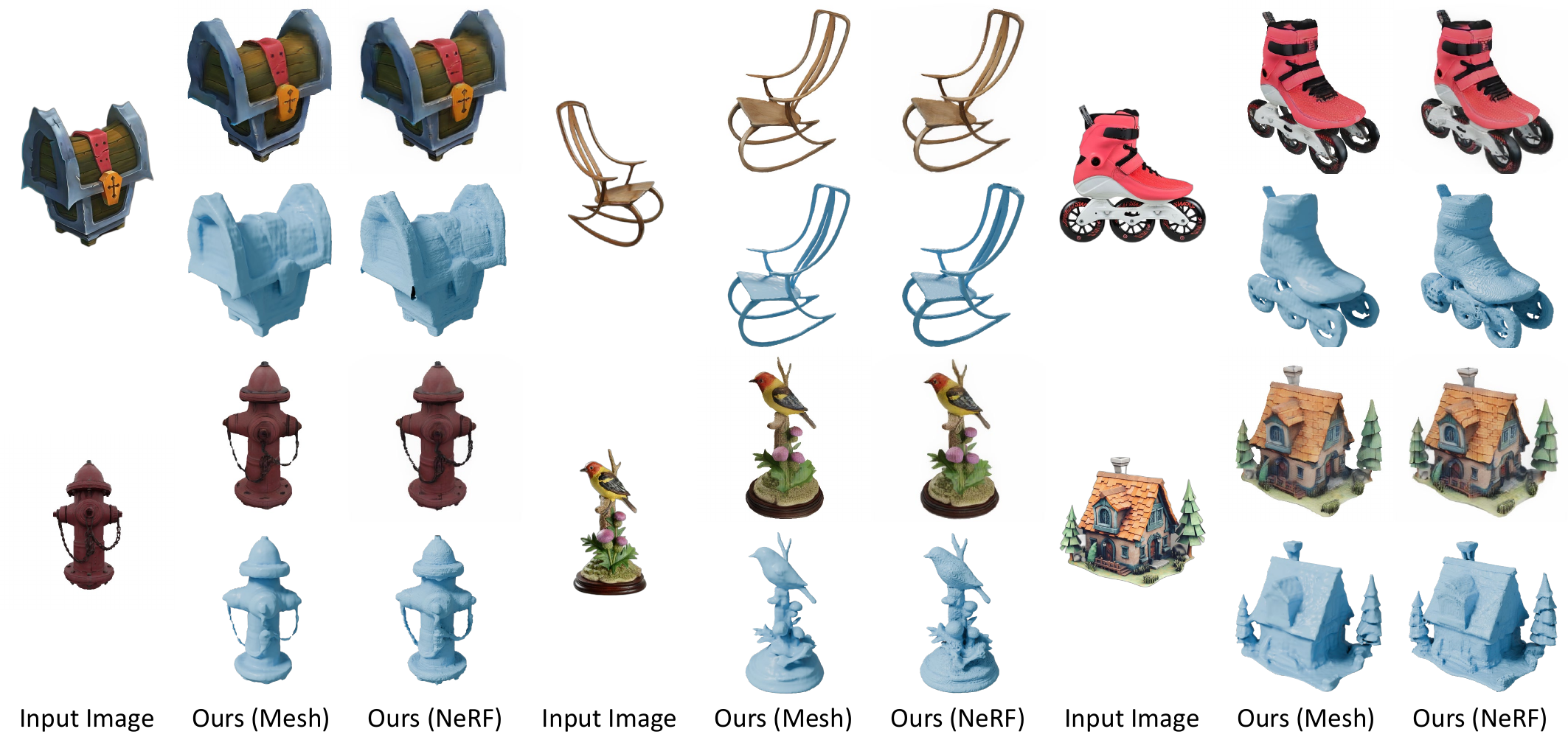}
\caption{Image-to-3D generation results using different sparse-view reconstruction model variants. For each generated mesh, we visualize both the textured rendering (upper) and untextured geometry (lower). All images are rendered at a fixed elevation of $20^{\circ}$.}
\label{fig:representation}
\end{figure*}

\noindent\textbf{Comparison between ``NeRF" and ``Mesh" variants.}
We also compare the ``Mesh'' and ``NeRF'' variants of our sparse-view reconstruction model quantitatively and qualitatively. From Table \ref{tab:gso}, \ref{tab:omni3d}, and \ref{tab:omni3dbm}, we can see that the ``NeRF" variant achieves slightly better metrics than the ``Mesh'' variant. We attribute this to the limited grid resolution of FlexiCubes, resulting in the lost of details when extracting mesh surfaces. Nevertheless, the drop in metrics is marginal and negligible considering the convenience brought by the efficient mesh rendering compared to the memory-intensive volume rendering of NeRF. Besides, we also visualize some image-to-3D generation results of the two model variants in Figure~\ref{fig:representation}. By applying explicit geometric supervisions, \ie, depths and normals, the ``Mesh'' model variant  can produce smoother surfaces compared to the meshes extracted from the density field of NeRF, which are generally more desirable in practical applications.

\section{Conclusion}
In this work, we present InstantMesh, an open-source instant image-to-3D framework that utilizes a transformer-based sparse-view large reconstruction model to create high-quality 3D assets from the images generated by a multi-view diffusion model. Building upon the Instant3D framework, we introduce mesh-based representation and additional geometric supervisions, significantly boosting the training efficiency and reconstruction quality. We also make improvements on other aspects, such as data preparation and training strategy. Evaluations on public datasets demonstrate that InstantMesh outperforms other latest image-to-3D baselines both qualitatively and quantitatively. InstantMesh is intended to make substantial contributions to the 3D Generative AI community and empower both researchers and creators.

\noindent\textbf{Limitations.} We notice that some limitations still exist in our framework and leave them for future work. (i) Following LRM~\cite{hong2024lrm} and Instant3D~\cite{li2024instantd}, our transformer-based triplane decoder produces $64\times 64$ triplanes, whose resolution may be a bottleneck for high-definition 3D modeling. (ii) Our 3D generation quality is inevitably influenced by the multi-view inconsistency of the diffusion model, while we believe this issue can be alleviated by utilizing more advanced multi-view diffusion architectures in the future. (iii) Although FlexiCubes can improve the smoothness and reduce the artifacts of the mesh surface due to the additional geometric supervisions, we notice that it is less effective on modeling tiny and thin structures compared to NeRF (Figure \ref{fig:representation}, \nth{2} row, \nth{1} column).

{
    \small
    \bibliographystyle{ieeenat_fullname}
    \bibliography{main}
}


\end{document}